\documentclass[11pt,a4paper]{article}
\usepackage[hyperref]{acl2021}
\usepackage{times}
\usepackage{latexsym}
\usepackage[pdftex]{graphicx}
\usepackage{amsmath}
\usepackage{multirow}
\usepackage{here}

\usepackage{microtype}

\aclfinalcopy 
\def\aclpaperid{11} 

\title{ARTA: Collection and Classification of Ambiguous Requests \\and Thoughtful Actions}

\author{
  Shohei Tanaka$^{1,2,3}$,
  Koichiro Yoshino$^{3,1,2}$,
  Katsuhito Sudoh$^{1,2}$,
  Satoshi Nakamura$^{1,2}$ \\
  \texttt{\{tanaka.shohei.tj7, sudoh, s-nakamura\}@is.naist.jp}, \\
  \texttt{koichiro.yoshino@riken.jp} \\
	$^1$ Nara Institute of Science and Technology (NAIST) \\
	$^2$ Center for Advanced Intelligence Project (AIP), RIKEN \\
	$^3$ Guardian Robot Project (GRP), R-IH, RIKEN
}

\date{}

\begin{document}
\maketitle
\begin{abstract}
Human-assisting systems such as dialogue systems must take thoughtful, appropriate actions not only for clear and unambiguous user requests, but also for ambiguous user requests, even if the users themselves are not aware of their potential requirements.
To construct such a dialogue agent, we collected a corpus and developed a model that classifies ambiguous user requests into corresponding system actions.
In order to collect a high-quality corpus, we asked workers to input antecedent user requests whose pre-defined actions could be regarded as thoughtful.
Although multiple actions could be identified as thoughtful for a single user request, annotating all combinations of user requests and system actions is impractical.
For this reason, we fully annotated only the test data and left the annotation of the training data incomplete.
In order to train the classification model on such training data, we applied the positive/unlabeled (PU) learning method, which assumes that only a part of the data is labeled with positive examples.
The experimental results show that the PU learning method achieved better performance than the general positive/negative (PN) learning method to classify thoughtful actions given an ambiguous user request.
\end{abstract}

\section{Introduction\label{sec:intro}}

Task-oriented dialogue systems satisfy user requests by using pre-defined system functions (Application Programming Interface (API) calls).
Natural language understanding, a module to bridge user requests and system API calls, is an important technology for spoken language applications such as smart speakers \cite{wu-etal-2019-transferable}.
\par
Although existing spoken dialogue systems assume that users give explicit requests to the system \cite{YOUNG2010150}, users may not always be able to define and verbalize the content and conditions of their own requests clearly \cite{yoshino-etal-2017-information}.
On the other hand, human concierges or guides can respond thoughtfully even when the users' requests are ambiguous.
For example, when a user says, ``I love the view here,'' they can respond, ``Shall I take a picture?''
If a dialogue agent can respond thoughtfully to a user who does not explicitly request a specific function, but has some potential request, the agent can provide effective user support in many cases.
We aim to develop such a system by collecting a corpus of user requests and thoughtful actions (responses) of the dialogue agent.
We also investigate whether the system responds thoughtfully to the user requests.
\par
The Wizard of Oz (WOZ) method, in which two subjects are assigned to play the roles of a user and a system, is a common method for collecting a user-system dialogue corpus \cite{budzianowski-etal-2018-multiwoz,kang-etal-2019-recommendation}.
However, in the collection of thoughtful dialogues, the WOZ method faces the following two problems.
First, even humans have difficulty responding thoughtfully to every ambiguous user request.
Second, since the system actions are constrained by its API calls, the collected actions sometimes are infeasible.
To solve these problems, we pre-defined 70 system actions and asked crowd workers to provide the antecedent requests for which each action could be regarded as thoughtful.
\par
We built a classification model to recognize single thoughtful system actions given the ambiguous user requests.
However, such ambiguous user requests can be regarded as antecedent requests of multiple system actions.
For example, if the function ``searching for fast food'' and the function ``searching for a cafe'' are invoked in action to the antecedent request ``I'm hungry,'' both are thoughtful actions.
Thus, we investigated whether the ambiguous user requests have other corresponding system actions in the 69 system actions other than the pre-defined system actions.
We isolated a portion of collected ambiguous user requests from the corpus and added additional annotation using crowdsourcing.
The results show that an average of 9.55 different actions to a single user request are regarded as thoughtful.
\par
Since annotating completely multi-class labels is difficult in actual data collection \cite{10.1007/978-3-319-10602-1_48}, we left the training data as incomplete data prepared as one-to-one user requests and system actions.
We defined a problem to train a model on the incompletely annotated data and tested on the completely annotated data\footnote{The dataset is available at\\\url{https://github.com/ahclab/arta\_corpus}.}.
In order to train the model on the incomplete training data, we applied the positive/unlabeled (PU) learning method \cite{elkan2008pu,CEVIKALP2020107164}, which assumes that some of the data are annotated as positive and the rest are not.
The experimental results show that the proposed classifier based on PU learning has higher classification performances than the conventional classifier, which is based on general positive/negative (PN) learning.


\section{Thoughtful System Action to Ambiguous User Request\label{sec:dataset}}

\begin{table}[t!]
	\begin{center}
		\small
		\begin{tabular}{l|p{5.8cm}}
		\hline
		Level & Definition \\
		\hline
		Q1 & The actual, but unexpressed request \\
		Q2 & The conscious, within-brain description of the request \\
		Q3 & The formal statement of the request \\
		Q4 & The request as presented to the dialogue agent \\
		\hline
		\end{tabular}
		\normalsize
	\end{center}
	\caption{Levels of ambiguity in requests (queries) \cite{taylor1962asking,taylor1968question}}
	\label{tab:taylor_level}
\end{table}

Existing task-oriented dialogue systems assume that user intentions are clarified and uttered in an explicit manner; however, 
users often do not know what they want to request.
User requests in such cases are ambiguous.
\citet{taylor1962asking,taylor1968question} categorizes user states in information search into four levels according to their clarity, as shown in Table \ref{tab:taylor_level}.
\par
Most of the existing task-oriented dialogue systems \cite{mem2seq,vanzo-bastianelli-lemon:2019:W19-59} convert explicit user requests (Q3) into machine readable expressions (Q4).
Future dialogue systems need to take appropriate actions even in situations such as Q1 and Q2, where the users are not able to clearly verbalize their requests \cite{yoshino-etal-2017-information}.
We used crowdsourcing to collect ambiguous user requests and link them to appropriate system actions.
This section describes the data collection.

\subsection{Corpus Collection\label{sec:collection}}

\begin{figure}[t]
	\begin{center}
	\includegraphics[width=7cm]{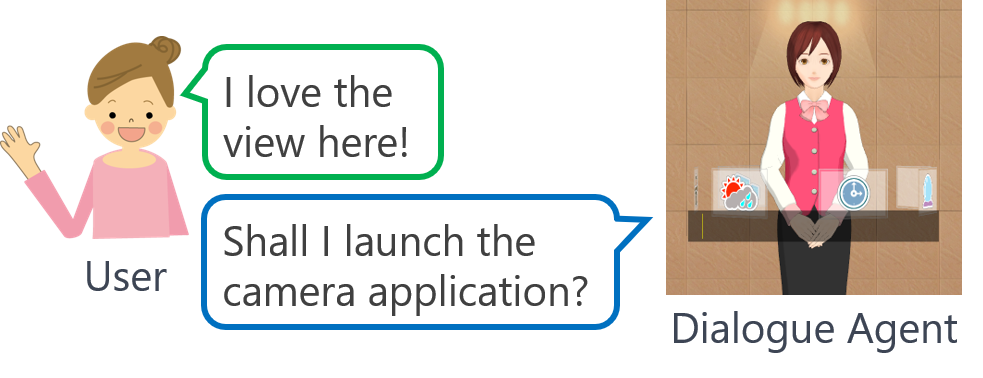}
	\end{center}
	\caption{Example of thoughtful dialogue}
	\label{fig:dialogue_example}
\end{figure}

We assume a dialogue between a user and a dialogue agent on a smartphone application in the domain of tourist information.
The user can make ambiguous requests or monologues, and the agent responds with thoughtful actions.
Figure \ref{fig:dialogue_example} shows an example dialogue between a user and a dialogue agent.
The user utterance ``I love the view here!'' is not verbalized as a request for a specific function.
The dialogue agent responds with a thoughtful action, ``Shall I launch the camera application?'' and launches the camera application.
\par
The WOZ method, in which two subjects are assigned to play the roles of a user and a dialogue agent, is widely used to collect dialogue samples.
However, even human workers have difficulty always responding with thoughtful actions to ambiguous user requests.
In other words, the general WOZ dialogue is not appropriate for collecting such thoughtful actions.
Moreover, these thoughtful actions must be linked to a system's API functions because possible agent actions are limited with its applications.
In other words, we can qualify the corpus by collecting antecedent ambiguous user requests to defined possible agent actions.
Therefore, we collected request-action pairs by asking crowd workers to input antecedent ambiguous user requests for the pre-defined agent action categories.
\par
We defined three major functions of the dialogue agent: ``spot search,'' ``restaurant search,'' and ``application (app) launch.''
Table \ref{tab:agent_category} shows the defined functions.
Each function has its own categories.
The actions of the dialogue agent in the corpus are generated by linking them to these categories.
There are 70 categories in total.
The functions and categories are defined heuristically according to Web sites for Kyoto sightseeing. 
``Spot search'' is a function to search for specific spots and is presented to the user in the form of an action such as ``Shall I search for an art museum around here?''
``Restaurant search'' is a function to search for specific restaurants and is presented to the user in the form of an action such as ``Shall I search for shaved ice around here?''
``App launch'' is a function to launch a specific application and is presented to the user in the form of an action such as ``Shall I launch the camera application?''
\par
We used crowdsourcing\footnote{https://crowdworks.jp/} to collect a Japanese corpus based on the pre-defined action categories of the dialogue agent\footnote{The details of the instruction and the input form are available in Appendix \ref{sec:instruction}.}.
The statistics of the collected corpus are shown in Table \ref{tab:corpus_statics}.
The request examples in the corpus are shown in Table \ref{tab:corpus_dialogue}.
Table \ref{tab:corpus_dialogue} shows that we collected ambiguous user requests where the pre-defined action could be regarded as thoughtful.
The collected corpus containing 27,230 user requests was split into training data:validation data:test data $ = 24,430:1,400:1,400$.
Each data set contains every category in the same proportion.

\begin{table}[t!]
	\begin{center}
		\small
		\begin{tabular}{p{1.2cm}|p{4.6cm}|r}
		\hline
		Function & Category & \# \\
		\hline
		spot search & amusement park, park, sports facility, experience-based facility, souvenir shop, zoo, aquarium, botanical garden, tourist information center, shopping mall, hot spring, temple, shrine, castle, nature or landscape, art museum, historic museum, kimono rental, red leaves, cherry blossom, rickshaw, station, bus stop, rest area, Wi-Fi spot, quiet place, beautiful place, fun place, wide place, nice view place & 30 \\
		\hline
		restaurant search & cafe, matcha, shaved ice, Japanese sweets, western-style sweets, curry, obanzai (traditional Kyoto food), tofu cuisine, bakery, fast food, noodles, nabe (Japanese stew), rice bowl or fried food, meat dishes, sushi or fish dishes, flour-based foods, Kyoto cuisine, Chinese, Italian, French, child-friendly restaurant or family restaurant, cha-kaiseki (tea-ceremony dishes), shojin (Japanese Buddhist vegetarian cuisine), vegetarian restaurant, izakaya or bar, food court, breakfast, inexpensive restaurant, average priced restaurant, expensive restaurant & 30 \\
		\hline
		app launch & camera, photo, weather, music, transfer navigation, message, phone, alarm, browser, map & 10 \\
		\hline
		\end{tabular}
		\normalsize
  \end{center}
  \caption{Functions and categories of dialogue agent. \# means the number of categories.}
	\label{tab:agent_category}
\end{table}

\begin{table*}[t!]
	\begin{center}
		\small
		\begin{tabular}{p{7.5cm}|p{7.5cm}}
		\hline
		User request (collecting with crowdsourcing) & System action (pre-defined) \\
		\hline
		I'm sweaty and uncomfortable. & Shall I search for a hot spring around here? \\
		I've been eating a lot of Japanese food lately and I'm getting a little bored of it. & Shall I search for meat dishes around here? \\
		Nice view. & Shall I launch the camera application? \\
		\hline
		\end{tabular}
		\normalsize
	\end{center}
	\caption{Examples of user requests in corpus. The texts are translated from Japanese to English. User requests for all pre-defined system actions are available in Appendix \ref{sec:additional}.}
	\label{tab:corpus_dialogue}
\end{table*}

\begin{table}[t!]
	\begin{center}
		\small
		\begin{tabular}{l|l|r}
		\hline
		Function & Ave. length & \# requests \\
		\hline
		spot search & \multicolumn{1}{r|}{$13.44\ (\pm4.69)$} & 11,670 \\
		restaurant search & \multicolumn{1}{r|}{$14.08\ (\pm4.82)$} & 11,670 \\
		app launch & \multicolumn{1}{r|}{$13.08\ (\pm4.65)$} & 3,890 \\
		\hline
		all & \multicolumn{1}{r|}{$13.66\ (\pm4.76)$} & 27,230 \\
		\hline
		\end{tabular}
		\normalsize
	\end{center}
	\caption{Corpus statistics}
	\label{tab:corpus_statics}
\end{table}

\subsection{Multi-Class Problem on Ambiguous User Request\label{sec:multilabel}}

Since the user requests collected in Sec. \ref{sec:collection} are ambiguous in terms of their requests, 
some of the 69 unannotated actions other than the pre-defined actions can be thoughtful.
Although labeling all combinations of user requests and system actions as thoughtful or not is costly and impractical, 
a comprehensive study is necessary to determine real thoughtful actions.
Thus, we completely annotated all combinations of 1,400 user requests and system actions in the test data.
\par
We used crowdsourcing for this additional annotation.
The crowd workers were presented with a pair of a user request and an unannotated action, and asked to make a binary judgment on whether the action was ``contextually natural and thoughtful to the user request'' or not.
Each pair was judged by three workers and the final decision was made by majority vote.
\par
The number of added action categories that were identified as thoughtful is shown in Table \ref{tab:multilabel}.
8.55 different categories on average were identified as thoughtful.
The standard deviation was 7.84; this indicates that the number of added categories varies greatly for each user request.
Comparing the number of added categories for each function, ``restaurant search'' has the highest average at 9.81 and ``app launch'' has the lowest average at 5.06.
The difference is caused by the target range of functions; ``restaurant search'' contains the same intention with different slots, while ``app launch'' covers different types of system roles.
For the second example showed in Table \ref{tab:corpus_dialogue}, ``I've been eating a lot of Japanese food lately, and I'm getting a little bored of it,'' suggesting any type of restaurant other than Japanese can be a thoughtful response in this dialogue context.
\par
Table \ref{tab:agreement} shows the detailed decision ratios of the additional annotation.
The ratios that two or three workers identified each pair of a user request and a system action as thoughtful are 7.23 and 5.16, respectively; this indicates that one worker identified about 60\% added action categories as not thoughtful.
Fleiss' kappa value is 0.4191; the inter-annotator agreement is moderate.
\par
\begin{figure}[t!]
	\begin{center}
	\includegraphics[width=7cm]{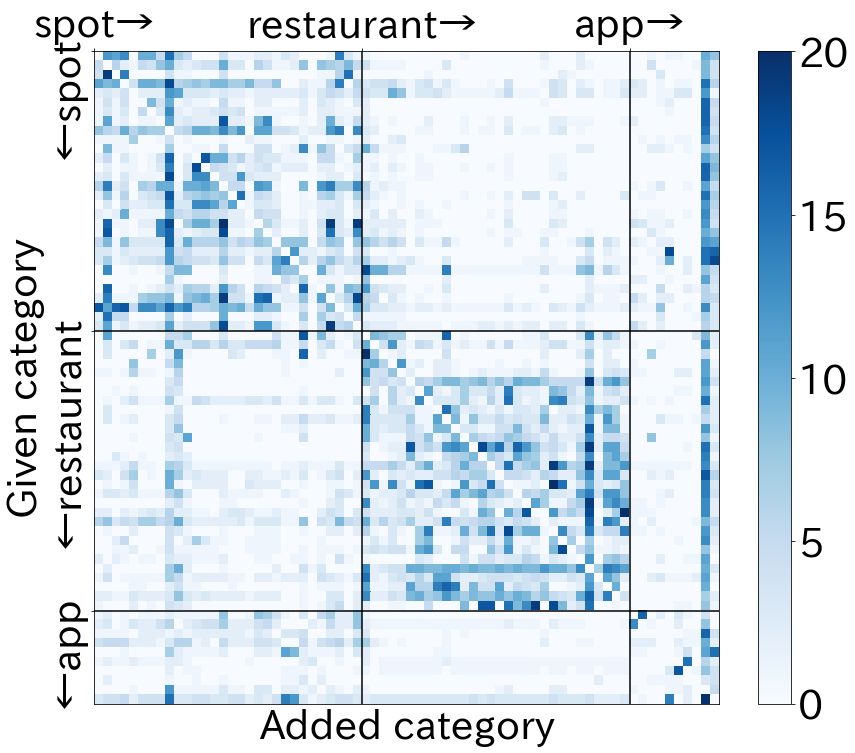}
	\end{center}
	\caption{Heat map of given and added categories}
	\label{fig:heatmap}
\end{figure}

Figure \ref{fig:heatmap} shows the heatmap of the given and added categories.
From the top left of both the vertical and horizontal axes, each line indicates one category in the order listed in Table \ref{tab:agent_category}.
The highest value corresponding to the darkest color in Figure \ref{fig:heatmap} is 20 because 20 ambiguous user requests are contained for each given action in the test data.
Actions related to the same role are annotated in functions of ``spot search'' and ``restaurant search.''
One of the actions near the rightmost column is identified as thoughtful for many contexts.
This action category was ``browser'' in the ``app launch'' function, which is expressed in the form of ``Shall I display the information about XX?''
``Spot search'' and ``restaurant search'' also had one action category annotated as thoughtful action for many antecedent requests.
These categories are, respectively, ``tourist information center'' and ``food court.''
\par
Table \ref{tab:heat_pair} shows some pairs that have large values in Fig. \ref{fig:heatmap}.
For any combination, both actions can be responses to the given ambiguous requests.

\begin{table}[t!]
	\begin{center}
		\small
		\begin{tabular}{l|r}
		\hline
		Function & \# added categories \\
		\hline
		spot search & $8.45\ (\pm7.34)$ \\
		restaurant search & $9.81\ (\pm7.77)$ \\
		app launch & $5.06\ (\pm8.48)$ \\
		\hline
		all & $8.55\ (\pm7.84)$ \\
		\hline
		\end{tabular}
		\normalsize
  \end{center}
  \caption{\# of added action categories}
	\label{tab:multilabel}
\end{table}

\begin{table}[t!]
	\begin{center}
		\small
		\begin{tabular}{l|r}
		\hline
		\# & Ratio (\%) \\
		\hline
		0 & $70,207\ (72.68)$ \\
		1 & $14,425\ (14.93)$ \\
		2 & $6,986\ (7.23)$ \\
		3 & $4,982\ (5.16)$ \\
		\hline
		all & $96,600$ \\
		\hline
		\end{tabular}
		\normalsize
  \end{center}
  \caption{Decision ratios of additional annotation. \# means the number of workers that identified each pair of a request and an action as thoughtful. The Fleiss' kappa value is 0.4191.}
	\label{tab:agreement}
\end{table}

\begin{table*}[t!]
	\begin{center}
		\small
		\begin{tabular}{l|l|r|l}
		\hline
		Pre-defined category & Added category & Frequency & Example user request \\
		\hline
		map & browser & 20 & Is XX within walking distance?\\
		red leaves & nature or landscape & 20 & I like somewhere that feels like autumn. \\
		shaved ice & cafe & 20 & I'm going to get heatstroke.\\
		French & expensive restaurant & 20 & I'm having a luxurious meal today!\\
		Kyoto cuisine & cha-kaiseki & 20 & I'd like to try some traditional Japanese food.\\
		\hline
		\end{tabular}
		\normalsize
  \end{center}
  \caption{Frequent pairs of pre-defined and additional categories. The user requests in Japanese are translated into English.}
	\label{tab:heat_pair}
\end{table*}


\section{Thoughtful Action Classification\label{sec:classifier}}

We collected pairs of ambiguous user requests and thoughtful system action categories in Sec. \ref{sec:dataset}.
Using this data, we developed a model that outputs thoughtful actions to given ambiguous user requests.
The model classifies user requests into categories of corresponding actions.
Positive/negative (PN) learning is widely used for classification, where the collected ambiguous user requests and the corresponding system action categories are taken as positive examples, and other combinations are taken as negative examples.
However, as indicated in Sec. \ref{sec:multilabel}, several action candidates can be thoughtful response actions to one ambiguous user request.
Since complete annotation to any possible system action is costly,
we apply positive/unlabeled (PU) learning to consider the data property; one action is annotated as a thoughtful response to one ambiguous user request, but labels of other system actions are not explicitly decided.
In this section, we describe the classifiers we used: a baseline system based on PN learning and the proposed system trained by the PU learning objective.

\subsection{Classifier}

\begin{figure}[tbp!]
	\begin{center}
	\includegraphics[width=7cm]{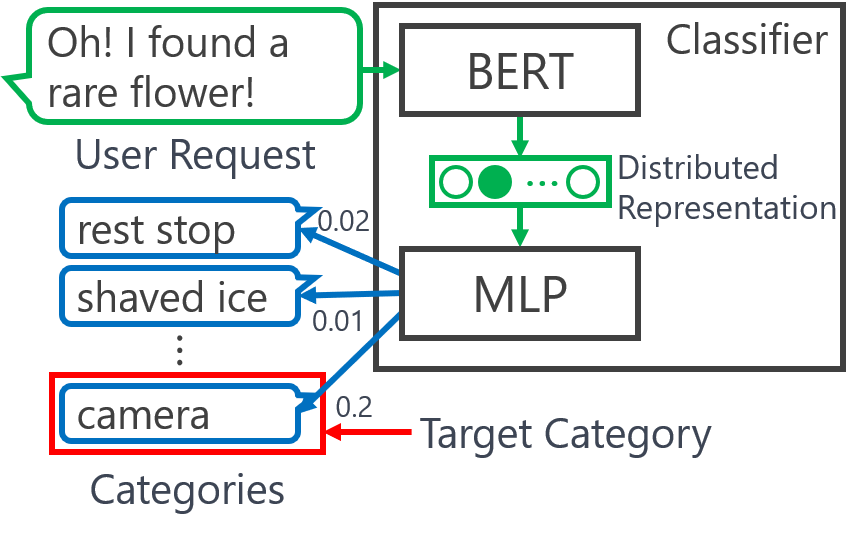}
	\end{center}
	\caption{User request classifier}
	\label{fig:classifier}
\end{figure}

Figure \ref{fig:classifier} shows the overview of the classification model.
The model classifies the ambiguous user requests into thoughtful action (positive example) categories of the dialogue agent.
We made a representation of a user request by Bidirectional Encoder Representations from Transformers (BERT) \cite{bert}, computed the mean vectors of the distributed representations given by BERT, and used them as inputs of a single-layer MultiLayer Perceptron (MLP).

\subsection{Loss Function in PN Learning\label{sec:loss_pn}}

When we simply build a classifier based on PN learning, the following loss function \cite{CEVIKALP2020107164} is used to train the model:
\small
\begin{eqnarray}
	Loss &=& \sum^{|U_{train}|}_{i} \sum^{|C^{+}_{x_i}|}_{j=1} \sum^{|C^{-}_{x_i}|}_{k=1} L(r_j) R_s (\mathbf{w}^{\mathsf{T}}_j \mathbf{x}_i - \mathbf{w}^{\mathsf{T}}_k \mathbf{x}_i) \nonumber \\
	&& + \kappa \sum^{|U_{train}|}_{i} \sum^{|C|}_{j=1} R_s (y_{ij}(\mathbf{w}^{\mathsf{T}}_j \mathbf{x}_i)). \label{eq:loss_pn}
\end{eqnarray}
\normalsize
\noindent
$U_{train}$ is the set of user requests included in the training data.
$C^{+}_{x_i}$ and $C^{-}_{x_i}$ are, respectively, the set of the positive example action categories associated with the user request $x_i$ and the set of the action categories without any annotation.
$r_j$ is the rank predicted by the model for the positive category $j$ and $L(r_j)$ is the weight function satisfying the following equation:
\begin{eqnarray}
	L(r) &=& \sum^r_{j=1} \frac{1}{j}. \label{eq:weight_l}
\end{eqnarray}
\noindent
Equation (\ref{eq:weight_l}) takes a larger value when the predicted rank is far from first place.
$\mathbf{w}_j$ is the weight vector corresponding to category $j$.
$\mathbf{x}_i$ is the distributed representation corresponding to user request $x_i$.
$R_s(t)$ is the ramp loss, which is expressed as,
\begin{eqnarray}
	R_s(t) &=& \min (1-m, \max(0, 1-t)). \label{eq:ramp}
\end{eqnarray}
\noindent
$m$ is a hyperparameter that determines the classification boundary.
Let $C$ be the set of defined categories, with $|C| = 70$.
$y_{ij}$ is $1$ if the category $j$ is a positive example for user request $x_i$ and $-1$ if it is not annotated.
$\kappa$ is a hyperparameter representing the weight of the second term.

\subsection{Loss Function in PU Learning\label{sec:loss_pu}}

Although the loss function of PN learning treats all combinations of unlabeled user requests and system action categories as negative examples, 
about 10\% of these combinations should be treated as positive examples in our corpus, as investigated in Sec. \ref{sec:multilabel}.
In order to consider the data property, we apply PU learning \cite{elkan2008pu}, which is an effective method for problems that are difficult to annotate completely, such as object recognition in images with various objects \cite{7780924}.
\par
We use a PU learning method proposed by \citet{CEVIKALP2020107164}, which is based on label propagation \cite{zhou2005learning,cevikalp2008semi}.
This method propagates labels of annotated samples to unlabeled samples using distance on a distributed representation space.
The original method \cite{CEVIKALP2020107164} propagates labels from the nearest neighbor samples on the distributed representation space.
The method calculates the similarity score $s_{ij}$ of the propagated labels (categories) as follows:
\begin{eqnarray}
	s_{ij} &=& \exp \left(- \frac{d(\mathbf{x}_i, \mathbf{x}_{j})}{\bar{d}} \cdot \frac{70}{69} \right) \label{eq:sim_nearest}.
\end{eqnarray}
\noindent
$\mathbf{x}_{j}$ is the vector of distributed representations of the nearest neighbor user request whose category $j$ is a positive example.
$d(\mathbf{x}_i, \mathbf{x}_{j})$ is the Euclidean distance between $\mathbf{x}_i$ and $\mathbf{x}_{j}$, and $\bar{d}$ is the mean of all distances.
The value range of $s_{ij}$ is $0\leq s_{ij}\leq 1$. It takes larger values when the Euclidean distance between two distributed representations becomes smaller.
We call this method {\bf (PU, nearest)}. 
\par
However, the original method is sensitive for outliers.
Thus, we propose a method to use the mean vectors of the user requests with the same category.
This method propagates labels according to their distance from these mean vectors.
We update the similarity score $s_{ij}$ in Eq. (\ref{eq:sim_nearest}) as follows:
\begin{eqnarray}
	s_{ij} &=& \exp \left(- \frac{d(\mathbf{x}_i, \bar{\mathbf{x}_{j}})}{\bar{d}} \cdot \frac{70}{69} \right).
\end{eqnarray}
\noindent
$\bar{\mathbf{x}_{j}}$ is the mean vector of distributed representations of the user requests whose category $j$ is a positive example.
We call this method {\bf (PU, mean)}.
The proposed method scales the similarity score $s_{ij}$ to a range of $-1\leq s_{ij}\leq 1$ using the following formula:
\begin{eqnarray}
	\tilde{s_{ij}} &=& -1 + \frac{2(s - \min(s))}{\max(s) - \min(s)} \label{eq:scale}.
\end{eqnarray}
\noindent
If the scaled score $\tilde{s_{ij}}$ is $0\leq \tilde{s_{ij}} \leq 1$, we add the category $j$ to $C^{+}_{x_i}$ and let $\tilde{s_{ij}}$ be the weight of category $j$ as a positive category.
If $\tilde{s_{ij}}$ is $-1\leq \tilde{s_{ij}} < 0$, category $j$ is assigned a negative label and the weight is set to $- \tilde{s_{ij}}$.
Using the similarity score $\tilde{s_{ij}}$, we update Eq. (\ref{eq:loss_pn}) as follows:
\small
\begin{eqnarray}
	&&Loss = \nonumber \\
	&&\sum^{|U_{train}|}_{i} \sum^{|C^{+}_{x_i}|}_{j=1} \sum^{|C^{-}_{x_i}|}_{k=1} \tilde{s_{ij}} \tilde{s_{ik}} L(r_j) R_s (\mathbf{w}^{\mathsf{T}}_j \mathbf{x}_i - \mathbf{w}^{\mathsf{T}}_k \mathbf{x}_i) \nonumber \\
	&&+ \kappa \sum^{|U_{train}|}_{i} \sum^{|C|}_{j=1} \tilde{s_{ij}} R_s (y_{ij}(\mathbf{w}^{\mathsf{T}}_j \mathbf{x}_i)). \label{eq:loss_pu}
\end{eqnarray}
\normalsize
In Eq. (\ref{eq:loss_pu}), $\tilde{s_{ij}}$ is a weight representing the contribution of the propagated category to the loss function.
The similarity score $\tilde{s_{ij}}$ of the annotated samples is set to 1.

\section{Experiments}

We evaluate the models developed in Sec. \ref{sec:classifier}, which classify user requests into the corresponding action categories.

\subsection{Model Configuration}

PyTorch \cite{NEURIPS2019_9015} is used to implement the models.
We used the Japanese BERT model \cite{kurohashi-bert}, which was pre-trained on Wikipedia articles.
Both BASE and LARGE model sizes \cite{bert} were used for the experiments.
\par
We used Adam \cite{adam} to optimize the model parameters and set the learning rate to $1\mathrm{e}{-5}$.
For $m$ in Eq. (\ref{eq:ramp}) and $\kappa$ in Eq. (\ref{eq:loss_pn}), we set $m = -0.8, \kappa = 5$ according to the literature \cite{CEVIKALP2020107164}.
We used the distributed representations output by BERT as the vector $\mathbf{x}_i$ in the label propagation.
Since the parameters of BERT are also optimized during the training, we reran the label propagation every five epochs.
We pre-trained the model by PN learning before we applied PU learning.
Similarity score $s_{ij}$ of {\bf (PU, nearest)} is also scaled by Eq. (\ref{eq:scale}) as with {\bf (PU, mean)}.
The parameters of each model used in the experiments were determined by the validation data.

\subsection{Evaluation Metrics}

Accuracy (Acc.), R@5 (Recall@5), and Mean Reciprocal Rank (MRR) were used as evaluation metrics.
R@5 counts the ratio of test samples, which have at least one correct answer category in their top five.
MRR ($ 0 < MRR \leq 1$) is calculated as follows:
\begin{eqnarray}
	MRR &=& \frac{1}{|U_{test}|} \sum^{|U_{test}|}_{i} \frac{1}{r_{x_i}}.
\end{eqnarray}
\noindent
$r_{x_i}$ means the rank output by the classification model for the correct answer category corresponding to user request $x_i$.
$U_{test}$ is the set of user requests included in the test data.
For all metrics, a higher value means better performance of the classification model.
The performance of each model was calculated from the average of ten trials.
For the test data, the correct action categories were annotated completely, as shown in Sec. \ref{sec:multilabel}; thus, multi-label scores were calculated for each model.

\subsection{Experimental Results}

\begin{table*}[t!]
	\begin{center}
		\begin{tabular}{l|r|r|r}
		\hline
		Model & Acc. (\%) & R@5 (\%) & MRR \\
		\hline
		BASE (PN) & $88.33\ (\pm0.92)$ & $97.99\ (\pm0.25)$ & $0.9255\ (\pm0.0056)$ \\
		BASE (PU, Nearest) & $88.29\ (\pm0.96)$ & $97.81\ (\pm0.27)$ & $0.9245\ (\pm0.0056)$ \\
		BASE (PU, Mean) & $^{\dagger}89.37\ (\pm0.78)$ & $97.85\ (\pm0.26)$ & $^{\dagger}0.9305\ (\pm0.0050)$ \\
		\hline
		LARGE (PN) & $89.16\ (\pm0.57)$ & $98.08\ (\pm0.22)$ & $0.9316\ (\pm0.0032)$ \\
		LARGE (PU, Nearest) & $89.06\ (\pm0.66)$ & $98.01\ (\pm0.24)$ & $0.9295\ (\pm0.0036)$ \\
		LARGE (PU, Mean) & $^{\dagger}90.13\ (\pm0.51)$ & $98.11\ (\pm0.27)$ & $^{\dagger}0.9354\ (\pm0.0035)$ \\
		\hline
		\end{tabular}
		\normalsize
	\end{center}
	\caption{Classification results. The results are the averages of ten trials.}
	\label{tab:multi_classification_results}
\end{table*}

\begin{table}[t!]
	\begin{center}
		\small
		\begin{tabular}{r|r|r}
		\hline
		Rank & Pre-defined category & \# Misclassifications \\
		\hline
		1 & browser & $6.95\ (\pm1.23)$ \\
		2 & average priced restaurant & $6.40\ (\pm1.50)$ \\
		3 & transfer navigation & $4.90\ (\pm1.02)$ \\
		4 & meat dishes & $4.35\ (\pm1.27)$ \\
		5 & park & $4.30\ (\pm1.30)$\\
		\hline
		\end{tabular}
		\normalsize
	\end{center}
	\caption{Frequent misclassification}
	\label{tab:false_count}
\end{table}

The experimental results are shown in Table \ref{tab:multi_classification_results}.
``PN'' is the scores of the PN learning method (Sec. \ref{sec:loss_pn}) and ``PU'' is the scores of the PU learning methods (Sec. \ref{sec:loss_pu}).
``Nearest'' means the label propagation considering only the nearest neighbor samples in the distributed representation space.
``Mean'' means the proposed label propagation using the mean vector of each category.
For each model, a paired t-test was used to test for significant differences in performance from the baseline (PN).
$\dagger$ means that $p < 0.01$ for a significant improvement in performance.
\par
Each system achieved more than 88 points for accuracy and 97 points for R@5.
The proposed method (PU, Mean) achieved significant improvement over the baseline method (PN); even the existing PU-based method (PU, Nearest) did not see this level of improvement.
We did not observe any improvements on R@5.
This probably means that most of the correct samples are already included in the top five, even in the baseline. 
We calculated the ratio of ``positive categories predicted by the PU learning model in the first place that are included in the positive categories predicted by the PN learning model in the second through fifth places'' when the following conditions were satisfied: ``the PN learning model does not predict any positive category in the first place,'' ``the PN learning model predicts some positive category in the second through fifth places,'' and ``the PU learning model predicts some positive category in the first place.''
The percentage is $95.53\ (\pm2.60)\%$, thus supporting our hypothesis for R@5.
\par
Table \ref{tab:false_count} shows the frequency of misclassification for each action category.
The number of misclassifications is calculated as the average of all models.
The results show that the most difficult category was ``browser,'' a common response category for any user request.

\subsection{Label Propagation Performance}

\begin{table*}[t!]
	\begin{center}
		\small
		\begin{tabular}{p{4cm}|p{3cm}|p{4cm}|p{3cm}}
		\hline
		Original request & Pre-defined category & Nearest request & Propagated category \\
		\hline
        I got some extra income today. & expensive restaurant & It's before payday. & inexpensive restaurant \\
		All the restaurants in the area seem to be expensive. & average priced restaurant & I want to try expensive ingredients. & expensive restaurant \\
		It's too rainy to go sightseeing. & fun place & I wonder when it's going to start raining today. & weather \\		
		\hline
		\end{tabular}
		\normalsize
	\end{center}
	\caption{Examples of wrong label propagations}
	\label{tab:wrong_lp}
\end{table*}

\begin{table}[t!]
	\begin{center}
		\begin{tabular}{l|r|r|r}
		\hline
		Model & Pre. (\%) & Rec. (\%) & F1 \\
		\hline
		\multirow{2}{*}{BASE} & $78.06$ & $8.53$ & $0.1533$ \\
		& $(\pm3.35)$ & $(\pm1.31)$ & $(\pm0.0206)$ \\
		\multirow{2}{*}{LARGE} & $79.27$ & $7.91$ & $0.1435$ \\
		& $(\pm4.43)$ & $(\pm1.10)$ & $(\pm0.0172)$ \\
		\hline
		\end{tabular}
		\normalsize
	\end{center}
	\caption{Label propagation performance}
	\label{tab:lp_performance}
\end{table}


\begin{table}[t!]
	\begin{center}
		\small
		\begin{tabular}{l|l|r}
		\hline
		Original & Propagated & Ratio (\%) \\
		\hline
		\multirow{3}{*}{spot search} & spot search & $16.71\ (\pm2.59)$ \\
		& restaurant search & $4.06\ (\pm1.27)$ \\
		& app launch & $6.81\ (\pm1.84)$ \\
		\hline
		\multirow{3}{*}{restaurant search} & spot search & $3.43\ (\pm1.01)$ \\
		& restaurant search & $43.06\ (\pm4.82)$ \\
		& app launch & $2.70\ (\pm0.64)$ \\
		\hline
		\multirow{3}{*}{app launch} & spot search & $10.94\ (\pm1.75)$ \\
		& restaurant search & $3.24\ (\pm1.13)$ \\
		& app launch & $9.06\ (\pm1.73)$ \\
		\hline
		\end{tabular}
		\normalsize
	\end{center}
	\caption{Ratios of false positive in label propagation}
	\label{tab:lp_fp_ratio}
\end{table}

In order to verify the effect of label propagation in PU learning, we evaluated the performance of the label propagation itself in the proposed method (PU, Mean) on the test data.
Table \ref{tab:lp_performance} shows the results.
Comparing Table \ref{tab:multi_classification_results} and Table \ref{tab:lp_performance}, the higher the precision of the label propagation, the higher the performance of the model.
For both models, more than 78\% of the propagated labels qualify as thoughtful.
We conclude that the label propagation is able to add thoughtful action categories as positive examples with high precision;
however, there is still room for improvement on their recalls.
\par
Table \ref{tab:wrong_lp} shows examples in which the label propagation failed.
``Nearest request'' is the nearest neighbor of ``original request'' among the requests labeled with ``propagated category'' as a positive example.
Comparing ``nearest request'' and ``original request'' in Table \ref{tab:wrong_lp}, the label propagation is mistaken when the sentence intentions are completely different or when the two requests contain similar words, but the sentence intentions are altered by negative forms or other factors.
\par
Table \ref{tab:lp_fp_ratio} shows the ratios of errors in the label propagation between the functions.
More than $40\%$ of the label propagation errors happened in the ``restaurant search'' category.
This is because the user request to eat is the same, but the narrowing down of the requested food is subject to subtle nuances, as shown in Table \ref{tab:wrong_lp}.


\section{Related Work}

We addressed the problem of building a natural language understanding system for ambiguous user requests, which is essential for task-oriented dialogue systems.
In this section, we discuss how our study differs from existing studies in terms of corpora for task-oriented dialogue systems and dealing with ambiguous user requests.

\subsection{Task-Oriented Dialogue Corpus\label{sec:corpus}}

Many dialogue corpora for task-oriented dialogue have been proposed, such as Frames \cite{el-asri-etal-2017-frames}, In-Car \cite{eric-etal-2017-key}, bAbI dialog \cite{DBLP:journals/corr/BordesW16}, and MultiWOZ \cite{budzianowski-etal-2018-multiwoz}.
These corpora assume that the user requests are clear, as in Q3 in Table \ref{tab:taylor_level} defined by \citet{taylor1962asking,taylor1968question}, and do not assume that user requests are ambiguous, as is the case in our study.
The corpus collected in our study assumes cases where the user requests are ambiguous, such as Q1 and Q2 in Table \ref{tab:taylor_level}.
\par
Some dialogue corpora are proposed to treat user requests that are not always clear: OpenDialKG \cite{moon-etal-2019-opendialkg}, ReDial \cite{NEURIPS2018_800de15c}, and RCG \cite{kang-etal-2019-recommendation}.
They assume that the system makes recommendations even if the user does not have a specific request, in particular, dialogue domains such as movies or music.
In our study, we focus on conversational utterance and monologue during sightseeing, which can be a trigger of thoughtful actions from the system.

\subsection{Disambiguation for User Requests}

User query disambiguation is also a conventional and important research issue in information retrieval \cite{di-marco-navigli-2013-clustering,wang-agichtein-2010-query,lee-etal-2002-implicit,towell-voorhees-1998-disambiguating}.
These studies mainly focus on problems of lexical variation, polysemy, and keyword estimation.
In contrast, our study focuses on cases where the user intentions are unclear.
\par
An interactive system to shape user intention is another research trend \cite{hixon-etal-2012-semantic,Guo2017LearningTQ}.
Such systems clarify user requests by interacting with the user with clarification questions.
\citet{bapna-etal-2017-sequential} collected a corpus and modeled the process with pre-defined dialogue acts.
These studies assume that the user has a clear goal request, while our system assumes that the user's intention is not clear.
In the corpus collected by \citet{cohen-lane-2012-simulation}, which assumes a car navigation dialogue agent, the agent responds to user requests classified as Q1, such as suggesting a stop at a gas station when the user is running out of gasoline.
Our study collected a variation of ambiguous user utterances to cover several situations in sightseeing.
\par
\citet{ohtake-etal-2009-annotating,yoshino-etal-2017-information} tackled sightseeing dialogue domains.
The corpus collected by \citet{ohtake-etal-2009-annotating} consisted of dialogues by a tourist and guide for making a one-day plan to sightsee in Kyoto.
However, it was difficult for the developed system to make particular recommendations for conversational utterances or monologues.
\citet{yoshino-etal-2017-information} developed a dialogue agent that presented information with a proactive dialogue strategy.
Although the situation is similar to our task,
their agent does not have clear natural language understanding (NLU) systems to bridge the user requests to a particular system action.


\section{Conclusion}

We collected a dialogue corpus that bridges ambiguous user requests to thoughtful system actions while focusing on system action functions (API calls).
We asked crowd workers to input antecedent user requests for which pre-defined dialogue agent actions could be regarded as thoughtful.
We also constructed test data as a multi-class classification problem, assuming cases in which multiple action candidates are qualified as thoughtful for the ambiguous user requests.
Furthermore, using the collected corpus, we developed classifiers that classify ambiguous user requests into corresponding categories of thoughtful system actions.
The proposed PU learning method achieved high accuracy on the test data, even when the model was trained on incomplete training data as the multi-class classification task.
\par
As future work, we will study the model architecture to improve classification performance.
It is particularly necessary to improve the performance of the label propagation.
We will also investigate the features of user requests that are difficult to classify.



\bibliographystyle{acl_natbib}
\bibliography{ref}

\appendix

\onecolumn 

\section{Appendix\label{sec:appendix}}

\subsection{Instruction and Input Form\label{sec:instruction}}

\begin{figure}[htbp!]
	\begin{center}
	\includegraphics[width=16cm]{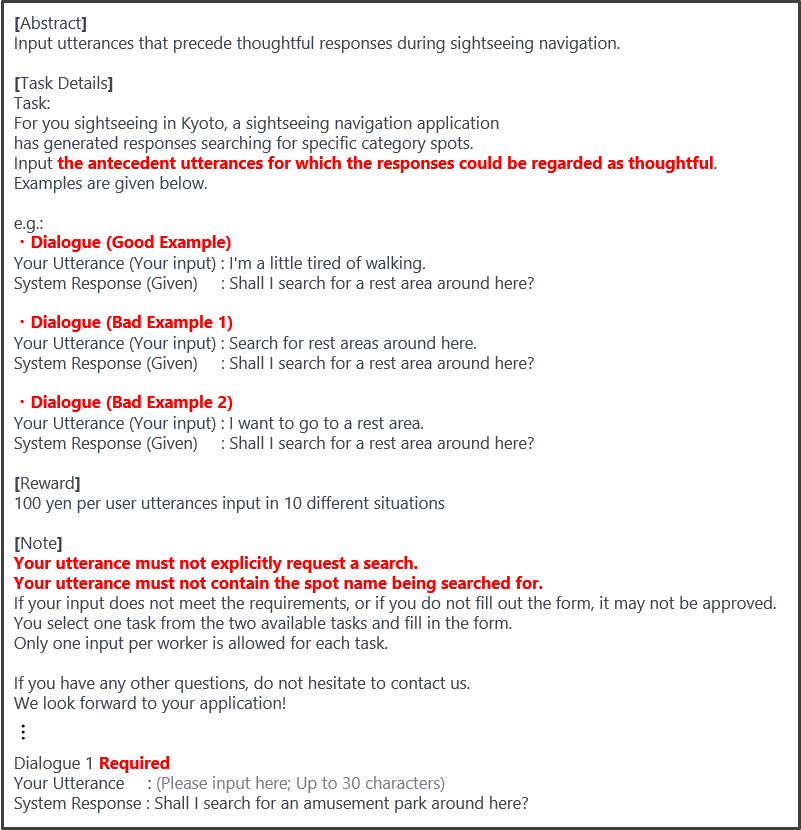}
	\end{center}
	\caption{Instruction and input form for corpus collection. The actual form is in Japanese; the figure is translated into English.}
	\label{fig:inst_cw}
\end{figure}

Figure \ref{fig:inst_cw} shows an example of an instruction and input form for the corpus collection.
Since the user requests (utterances) to be collected in our study need to be ambiguous, a bad example is an utterance with a clear request, such as, ``Search for rest areas around here.''
Each worker was asked to input user requests for ten different categories.

\subsection{Additional Examples of User Requests\label{sec:additional}}

\begin{table*}[t!]
	\begin{center}
		\scriptsize
		\begin{tabular}{p{7.5cm}|p{7.5cm}}
		\hline
		User request (collecting with crowdsourcing) & System action (pre-defined) \\
		\hline
		Is there a place where we can have fun as a family for a day? & Shall I search for an amusement park around here? \\
		I want to take a nap on the grass. & Shall I search for a park around here? \\
		I want to move my body as much as I can. & Shall I search for a sports facility around here? \\
		I'd like to do something more than just watch. & Shall I search for an experience-based facility around here? \\
		I want a Kyoto-style key chain. & Shall I search for a souvenir shop around here? \\
		Where can I see pandas? & Shall I search for a zoo around here? \\
		I haven't seen any penguins lately. & Shall I search for an aquarium around here? \\
		I want to relax in nature. & Shall I search for a botanical garden around here? \\
		I don't know where to go. & Shall I search for a tourist information center around here? \\
		It's suddenly getting cold. I need a jacket. & Shall I search for a shopping mall around here? \\
		I'm sweaty and uncomfortable. & Shall I search for a hot spring around here? \\
		I'm interested in historical places. & Shall I search for a temple around here? \\
		This year has not been a good one. & Shall I search for a shrine around here? \\
		I wonder if there are any famous buildings. & Shall I search for a castle around here? \\
		I need some healing. & Shall I search for nature or landscapes around here? \\
		It's autumn and it's nice to experience art. & Shall I search for an art museum around here? \\
		Is there a tourist spot where I can study as well? & Shall I search for an historic museum around here? \\
		I'd love to walk around a place like here wearing a kimono. & Shall I search for a kimono rental shop around here? \\
		I'd like to see some autumnal scenery. & Shall I search for red leaves around here? \\
		I want to feel spring. & Shall I search for cherry blossoms around here? \\
		I want to go on an interesting ride. & Shall I search for a rickshaw around here? \\
		It would be faster to go by train. & Shall I search for a station around here? \\
		It takes time on foot. & Shall I search for a bus stop around here? \\
		I'd like to sit down and relax. & Shall I search for a rest area around here? \\
		I'm having trouble getting good reception. & Shall I search for a WiFi spot around here? \\
		I want to relax. & Shall I search for a quiet place around here? \\
		I'd like to take a picture to remember the day. & Shall I search for a beautiful place around here? \\
		I wonder if there are any places where children can play. & Shall I search for a fun place around here? \\
		I want to feel liberated. & Shall I search for a wide place around here? \\
		I want to see the night view. & Shall I search for a place with a nice view around here? \\
		\hline
		I'm thirsty. & Shall I search for a cafe around here? \\
		I bought some delicious Japanese sweets! & Shall I search for matcha around here? \\
		It's so hot, I'm sweating all over. & Shall I search for shaved ice around here? \\
		I'm getting bored with cake. & Shall I search for Japanese sweets around here? \\
		I feel like having a 3 o'clock snack. & Shall I search for western-style sweets around here? \\
		I want something spicy! & Shall I search for curry around here? \\
		I'd like to eat something homey. & Shall I search for obanzai around here? \\
		I want to eat something healthy. & Shall I search for tofu cuisine around here? \\
		I want to buy some breakfast for tomorrow. & Shall I search for a bakery around here? \\
		I think it's time for a snack. & Shall I search for fast food around here? \\
		I'm not really in the mood for rice. & Shall I search for noodles around here? \\
		It's cold today, so I'd like to eat something that will warm me up. & Shall I search for nabe around here? \\
		I want to eat a heavy meal. & Shall I search for rice bowls or fried food around here? \\
		I've been eating a lot of Japanese food lately, and I'm getting a little bored of it. & Shall I search for meat dishes around here? \\
		I think I've been eating a lot of meat lately. & Shall I search for sushi or fish dishes around here? \\
		Let's have a nice meal together. & Shall I search for flour-based foods around here? \\
		I want to eat something typical of Kyoto. & Shall I search for Kyoto cuisine around here? \\
		My daughter wants to eat fried rice. & Shall I search for Chinese food around here? \\
		I'm not in the mood for Japanese or Chinese food today. & Shall I search for Italian food around here? \\
		It's a special day. & Shall I search for French food around here? \\
		The kids are hungry and whining. & Shall I search for a child-friendly restaurant or family restaurant around here? \\
		I wonder if there is a calm restaurant. & Shall I search for cha-kaiseki around here? \\
		I want to lose weight. & Shall I search for shojin around here? \\
		I hear the vegetables are delicious around here. & Shall I search for a vegetarian restaurant around here? \\
		It's nice to have a night out drinking in Kyoto! & Shall I search for an izakaya or bar around here? \\
		There are so many things I want to eat, it's hard to decide. & Shall I search for a food court around here? \\
		When I travel, I get hungry from the morning. & Shall I search for breakfast around here? \\
		I don't have much money right now. & Shall I search for an inexpensive restaurant around here? \\
		I'd like a reasonably priced restaurant. & Shall I search for an average priced restaurant around here? \\
		I'd like to have a luxurious meal. & Shall I search for an expensive restaurant around here? \\
		\hline
		Nice view. & Shall I launch the camera application? \\
		What did I photograph today? & Shall I launch the photo application? \\
		I hope it's sunny tomorrow. & Shall I launch the weather application? \\
		I want to get excited. & Shall I launch the music application? \\
		I'm worried about catching the next train. & Shall I launch the transfer navigation application? \\
		I have to tell my friends my hotel room number. & Shall I launch the message application? \\
		I wonder if XX is back yet. & Shall I call XX? \\
		The appointment is at XX. & Shall I set an alarm for XX o'clock? \\
		I wonder what events are going on at XX right now. & Shall I display the information about XX? \\
		How do we get to XX? & Shall I search for a route to XX? \\
		\hline
		\end{tabular}
		\normalsize
	\end{center}
	\caption{User requests for all pre-defined system actions. The texts are translated from Japanese to English.}
	\label{tab:additional_examples}
\end{table*}

Table \ref{tab:additional_examples} shows examples of user requests for all pre-defined system actions.

\end{document}